\documentclass[]{article}

\usepackage{url}
\usepackage{graphicx} 
\usepackage{algorithm}
\usepackage{algorithmic}
\usepackage{amsthm}
\usepackage{amsmath}
\usepackage{amsfonts}
\usepackage{bbm}
\usepackage{titlesec}
\usepackage{caption}
\usepackage{subcaption}
\usepackage{natbib}
\DeclareMathOperator{\softmax}{softmax}

\begin{document}

\title{A Critical Review of Recurrent Neural Networks for Sequence Learning}

\author{
  Zachary C. Lipton\\
  \texttt{zlipton@cs.ucsd.edu}
  \and
  John Berkowitz\\
  \texttt{jaberkow@physics.ucsd.edu}
  \and
  Charles Elkan\\
  \texttt{elkan@cs.ucsd.edu}}


\date{June 5th, 2015}
\maketitle

\begin{abstract}
Countless learning tasks require dealing with sequential data. 
Image captioning, speech synthesis, and music generation 
all require that a model produce outputs that are sequences.
In other domains, such as time series prediction, video analysis, and musical information retrieval,
a model must learn from inputs that are sequences.
Interactive tasks, such as translating natural language,
engaging in dialogue, and controlling a robot, 
often demand both capabilities.
Recurrent neural networks (RNNs) are connectionist models 
that capture the dynamics of sequences via cycles in the network of nodes.
Unlike standard feedforward neural networks, recurrent networks retain a state
that can represent information from an arbitrarily long context window.
Although recurrent neural networks have traditionally been difficult to train, and often contain millions of parameters,
recent advances in network architectures, optimization techniques, and parallel computation
have enabled successful large-scale learning with them.
In recent years, systems based on long short-term memory (LSTM)
and bidirectional (BRNN) architectures
have demonstrated ground-breaking performance on tasks as varied 
as image captioning, language translation, and handwriting recognition.
In this survey, we review and synthesize the research
that over the past three decades first yielded and then made practical these powerful learning models.
When appropriate, we reconcile conflicting notation and nomenclature.
Our goal is to provide a self-contained explication of the state of the art
together with a historical perspective and references to primary research.
\end{abstract}

\section{Introduction}
Neural networks are powerful learning models 
that achieve state-of-the-art results in a wide range of supervised and unsupervised machine learning tasks.
They are suited especially well for machine perception tasks, 
where the raw underlying features are not individually interpretable. 
This success is attributed to their ability to learn hierarchical representations,
unlike traditional methods that rely upon hand-engineered features \citep{farabet2013learning}.
Over the past several years,  storage has become more affordable, datasets have grown far larger, 
and the field of parallel computing has advanced considerably.
In the setting of large datasets, simple linear models tend to under-fit,
and often under-utilize computing resources.
Deep learning methods, in particular those based on {deep belief networks} (DNNs), 
which are greedily built by stacking restricted Boltzmann machines,
and convolutional neural networks, 
which exploit the local dependency of visual information, 
have demonstrated record-setting results on many important applications.

However, despite their power, 
standard neural networks have limitations. 
Most notably, they rely on the assumption of independence among the training and test examples.
After each example (data point) is processed, the entire state of the network is lost. 
If each example is generated independently, this presents no problem.
But if data points are related in time or space, this is unacceptable.
Frames from video, snippets of audio, and words pulled from sentences,  
represent settings where the independence assumption fails. 
Additionally, standard networks generally rely on examples being vectors of fixed length.
Thus it is desirable to extend these powerful learning tools 
to model data with temporal or sequential structure and varying length inputs and outputs,
especially in the many domains where neural networks are already the state of the art.
Recurrent neural networks (RNNs) are connectionist models
with the ability to selectively pass information across sequence steps, 
while processing sequential data one element at a time.
Thus they can model input and/or output
consisting of sequences of elements that are not independent. 
Further, recurrent neural networks can simultaneously model sequential and time dependencies 
on multiple scales. 

In the following subsections, we explain the fundamental reasons 
why recurrent neural networks are worth investigating.
To be clear, we are motivated by a desire to achieve empirical results. 
This motivation warrants clarification because recurrent networks have roots 
in both cognitive modeling and supervised machine learning.
Owing to this difference of perspectives, many published papers have different aims and priorities.
In many foundational papers, generally published in cognitive science and computational neuroscience journals,
such as \citep{hopfield1982neural, jordan1997serial, elman1990finding}, 
biologically plausible mechanisms are emphasized.
In other papers \citep{schuster1997bidirectional, socher2014grounded, karpathy2014deep},
biological inspiration is downplayed in favor of achieving empirical results 
on important tasks and datasets.
This review is motivated by practical results rather than biological plausibility, 
but where appropriate, we draw connections to relevant concepts in neuroscience.
Given the empirical aim, we now address three significant questions 
that one might reasonably want answered before reading further. 

\subsection{Why model sequentiality explicitly?}

In light of the practical success and economic value of sequence-agnostic models,
this is a fair question.
Support vector machines, logistic regression, and feedforward networks 
have proved immensely useful without explicitly modeling time. 
Arguably, it is precisely the assumption of independence 
that has led to much recent progress in machine learning.
Further, many models implicitly capture time by concatenating each input 
with some number of its immediate predecessors and successors,
presenting the machine learning model 
with a sliding window of context about each point of interest. 
This approach has been used with deep belief nets for speech modeling by \citet{maas2012recurrent}.

Unfortunately, despite the usefulness of the independence assumption,
it precludes modeling long-range dependencies.
For example, a model trained using a finite-length context window of length $5$
could never be trained to answer the simple question,
\emph{``what was the data point seen six time steps ago?"}
For a practical application such as call center automation, 
such a limited system might learn to route calls,
but could never participate with complete success in an extended dialogue.
Since the earliest conception of artificial intelligence, 
researchers have sought to build systems that interact with humans in time. 
In Alan Turing's groundbreaking paper \emph{Computing Machinery and Intelligence}, 
he proposes an ``imitation game" which judges a machine's intelligence by
its ability to convincingly engage in dialogue \citep{turing1950computing}.
Besides dialogue systems, modern interactive systems of economic importance include
self-driving cars and robotic surgery, among others.
Without an explicit model of sequentiality or time, 
it seems unlikely that any combination of classifiers or regressors 
can be cobbled together to provide this functionality.

\subsection{Why not use Markov models?}

Recurrent neural networks are not the only models capable of representing time dependencies. 
Markov chains, which model transitions between states in an observed sequence, 
were first described by the mathematician Andrey Markov in 1906. 
Hidden Markov models (HMMs), which model an observed sequence
as probabilistically dependent upon a sequence of unobserved states, 
were described in the 1950s and have been widely studied since the 1960s \citep{stratonovich1960conditional}.
However, traditional Markov model approaches are limited because their states 
must be drawn from a modestly sized discrete state space $S$.
The dynamic programming algorithm that is used to perform efficient inference with hidden Markov models 
scales in time $O(|S|^2)$ \citep{viterbi1967error}. 
Further, the transition table capturing the probability of moving between any two time-adjacent states is of size $|S|^2$.
Thus, standard operations become infeasible with an HMM 
when the set of possible hidden states grows large.
Further, each hidden state can depend only on the immediately previous state.
While it is possible to extend a Markov model 
to account for a larger context window 
by  creating a new state space equal to the cross product 
of the possible states at each time in the window, 
this procedure grows the state space exponentially with the size of the window,
rendering Markov models computationally impractical for modeling long-range dependencies \citep{graves2014neural}.

Given the limitations of Markov models, 
we ought to explain why it is reasonable that connectionist models, 
i.e., artificial neural networks, should fare better.
First, recurrent neural networks can capture long-range time dependencies, 
overcoming the chief limitation of Markov models.
This point requires a careful explanation. 
As in Markov models, any state in a traditional RNN depends only on the current input 
as well as on the state of the network at the previous time step.\footnote{
While traditional RNNs only model the dependence of the current state on the previous state,
bidirectional recurrent neural networks (BRNNs) \citep{schuster1997bidirectional} extend RNNs
to model dependence on both past states and future states. 
}
However, the hidden state at any time step can contain information 
from a nearly arbitrarily long context window. 
This is possible because the number of distinct states
that can be represented in a hidden layer of nodes
grows exponentially with the number of nodes in the layer.
Even if each node took only binary values, 
the network could represent $2^N$ states
where $N$ is the number of nodes in the hidden layer.
When the value of each node is a real number,
a network can represent even more 
distinct states.
While the potential expressive power of a network grows exponentially with the number of nodes,
the complexity of both inference and training grows at most quadratically.

\subsection{Are RNNs too expressive?}

Finite-sized RNNs with nonlinear activations are a rich family of models, 
capable of nearly arbitrary computation. 
A well-known result 
is that a finite-sized recurrent neural network with sigmoidal activation functions
can simulate a universal Turing machine  \citep{siegelmann1991turing}.
The capability of RNNs to perform arbitrary computation
demonstrates their expressive power,
but one could argue that the C programming language 
is equally capable of expressing arbitrary programs. 
And yet there are no papers claiming that the invention of C 
represents a panacea for machine learning.
A fundamental reason is there is no simple way of efficiently exploring the space of C programs.
In particular, there is no general way to calculate the gradient of an arbitrary C program 
to minimize a chosen loss function.
Moreover, given any finite dataset, 
there exist countless programs which overfit the dataset, 
generating desired training output but failing to generalize to test examples.

Why then should RNNs suffer less from similar problems? 
First, given any fixed architecture (set of nodes, edges, and activation functions), 
the recurrent neural networks with this architecture are differentiable end to end. 
The derivative of the loss function can be calculated with respect 
to each of the parameters (weights) in the model.
Thus, RNNs are amenable to gradient-based training.
Second, while the Turing-completeness of RNNs is an impressive property, 
given a fixed-size RNN with a specific architecture, 
it is not actually possible to reproduce any arbitrary program. 
Further, unlike a program composed in C, 
a recurrent neural network can be regularized via standard techniques 
that help prevent overfitting,
such as weight decay, dropout, and limiting the degrees of freedom. 

\subsection{Comparison to prior literature}

The literature on recurrent neural networks can seem impenetrable to the uninitiated.
Shorter papers assume familiarity with a large body of background literature,
while diagrams are frequently underspecified, 
failing to indicate which edges span time steps and which do not.
Jargon abounds, and notation is inconsistent across papers 
or overloaded within one paper.
Readers are frequently in the unenviable position 
of having to synthesize conflicting information across many papers in order to understand just one. 
For example, in many papers subscripts index both nodes and time steps.
In others, $h$ simultaneously stands for a link function and a layer of hidden nodes.
The variable $t$ simultaneously stands for both time indices and targets, sometimes in the same equation. 
Many excellent research papers have appeared recently, 
but clear reviews of the recurrent neural network literature are rare.

Among the most useful resources are 
a recent book on supervised sequence labeling 
with recurrent neural networks \citep{graves2012supervised} 
and an earlier doctoral thesis \citep{gers2001long}.
A recent survey covers recurrent neural nets for language modeling \citep{de2015survey}.
Various authors focus on specific technical aspects;
for example \citet{pearlmutter1995gradient} surveys gradient calculations in continuous time recurrent neural networks.
In the present review paper, we aim to provide a readable, intuitive, consistently notated,
and reasonably comprehensive but selective survey of research 
on recurrent neural networks for learning with sequences.
We emphasize architectures, algorithms, and results, 
but we aim also to distill the intuitions 
that have guided this largely heuristic and empirical field.
In addition to concrete modeling details, 
we offer qualitative arguments, a historical perspective,
and comparisons to alternative methodologies where appropriate.

\section{Background}

This section introduces formal notation and provides a brief background on neural networks in general.

\subsection{Sequences}

The input to an RNN is a sequence, and/or its target is a sequence.
An input sequence can be denoted 
$(\boldsymbol{x}^{(1)}, \boldsymbol{x}^{(2)}, ... , \boldsymbol{x}^{(T)})$
where each data point $\boldsymbol{x}^{(t)}$ is a real-valued vector.
Similarly, a target sequence can be denoted $(\boldsymbol{y}^{(1)}, \boldsymbol{y}^{(2)}, ... , \boldsymbol{y}^{(T)})$.
A training set typically is a set of examples where each example is an (input sequence, target sequence) pair,
although commonly either the input or the output may be a single data point. 
Sequences may be of finite or countably infinite length.
When they are finite, the maximum time index of the sequence is called $T$.
RNNs are not limited to time-based sequences.
They have been used successfully on non-temporal sequence data, 
including genetic data \citep{baldi2003principled}.
However, 
in many important applications of RNNs, 
the sequences have an explicit or implicit temporal aspect.
While we often refer to time in this survey,
the methods described here are applicable to non-temporal as well as to temporal tasks. 

Using temporal terminology, an input sequence consists of
data points $\boldsymbol{x}^{(t)}$ that arrive 
in a discrete {sequence} of \emph{time steps} indexed by $t$.
A target sequence consists of data points $\boldsymbol{y}^{(t)}$.
We use superscripts with parentheses for time, and not subscripts,
to prevent confusion between sequence steps and indices of nodes in a network.
When a model produces predicted data points, these are labeled $\hat{\boldsymbol{y}}^{(t)}$.

The time-indexed data points may be equally spaced samples from a continuous real-world process.
Examples include the still images that comprise the frames of videos
or the discrete amplitudes sampled at fixed intervals that comprise audio recordings.
The time steps may also be ordinal, with no exact correspondence to durations.
In fact, RNNs are frequently applied to domains 
where sequences have a defined order but no explicit notion of time.
This is the case with natural language. 
In the word sequence ``John Coltrane plays the saxophone", 
$\boldsymbol{x}^{(1)} = \text{John}$, $\boldsymbol{x}^{(2)} = \text{Coltrane}$, etc.

\subsection{Neural networks}

Neural networks are biologically inspired models of computation.
Generally, a neural network consists of a set of \emph{artificial neurons}, 
commonly referred to as \emph{nodes} or \emph{units},
and a set of directed edges between them, 
which intuitively represent the \emph{synapses}
in a biological neural network.
Associated with each neuron $j$ is an activation function $l_j(\cdot)$,
which is sometimes called a link function. 
We use the notation $l_j$ and not $h_j$, unlike some other papers,
to distinguish the activation function from the values of the hidden nodes in a network, 
which, as a vector, is commonly notated $\boldsymbol{h}$ in the literature.

Associated with each edge from node $j'$ to $j$ is a weight $w_{jj'}$.
Following the convention adopted in several foundational papers 
\citep{hochreiter1997long, gers2000learning, gers2001long, sutskever2011generating},
we index neurons with $j$ and $j'$,
and  $w_{jj'}$ denotes the ``to-from" weight corresponding to the directed edge to node $j$ from node $j'$.
It is important to note that in many references
the indices are flipped and $w_{j'j} \neq w_{jj'}$ 
denotes the ``from-to" weight on the directed edge from the node $j'$ to the node $j$,
as in lecture notes by \citet{elkanlearningmeaning} and in \citet{wiki:backpropagation}.

\begin{figure}
  \centering
  \includegraphics[width=.6\linewidth]{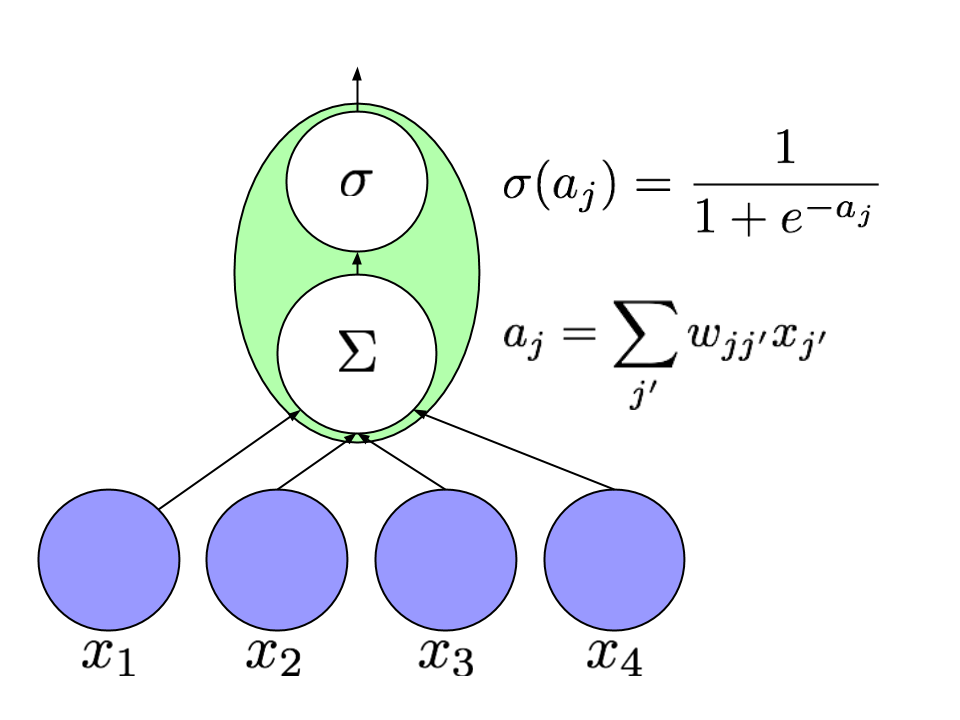}
  \caption{An artificial neuron computes a nonlinear function of a weighted sum of its inputs.}
  \label{fig:artificial-neuron}
\end{figure}%

The value $v_j$ of each neuron $j$ is calculated 
by applying its activation function to a weighted sum of the values of its input nodes (Figure~\ref{fig:artificial-neuron}):
$$v_j = l_j \left( \sum_{j'} w_{jj'} \cdot v_{j'} \right) .$$
For convenience, we term the weighted sum inside the parentheses 
the \emph{incoming activation} and notate it as $a_j$.
We represent this computation in diagrams by depicting  
neurons as circles and edges as arrows connecting them.
When appropriate, we indicate the exact activation function with a symbol, e.g., $\sigma$ for sigmoid.

Common choices for the activation function include the sigmoid $\sigma(z) = 1/(1 + e^{-z})$ 
and the $\textit{tanh}$ function $\phi(z) = (e^{z} - e^{-z})/(e^{z} + e^{-z})$.
The latter has become common in feedforward neural nets 
and was applied to recurrent nets by \citet{sutskever2011generating}.
Another activation function which has become prominent in deep learning research
is the rectified linear unit (ReLU) whose formula is $l_j(z) = \max(0,z)$. 
This type of unit has been demonstrated to improve the performance of many deep neural networks 
\citep{nair2010rectified, maas2012recurrent, zeiler2013rectified}
on tasks as varied as speech processing and object recognition,
and has been used in recurrent neural networks by \citet{bengio2013advances}.

The activation function at the output nodes depends upon the task. 
For multiclass classification with $K$ alternative classes, 
we apply a $\softmax$ nonlinearity in an output layer of $K$ nodes.
The $\softmax$ function calculates
$$
\hat{y}_k = \frac{e^{a_k} }{ \sum_{k'=1}^{K} e^{ {a_{k'}}  }} \mbox{~for~} k=1 \mbox{~to~} k=K.
$$
The denominator is a normalizing term consisting of the sum of the numerators,
ensuring that the outputs of all nodes sum to one.
For multilabel classification the activation function is simply a point-wise sigmoid,
and for regression we typically have linear output.

\begin{figure}
  \centering
  \includegraphics[width=.7\linewidth]{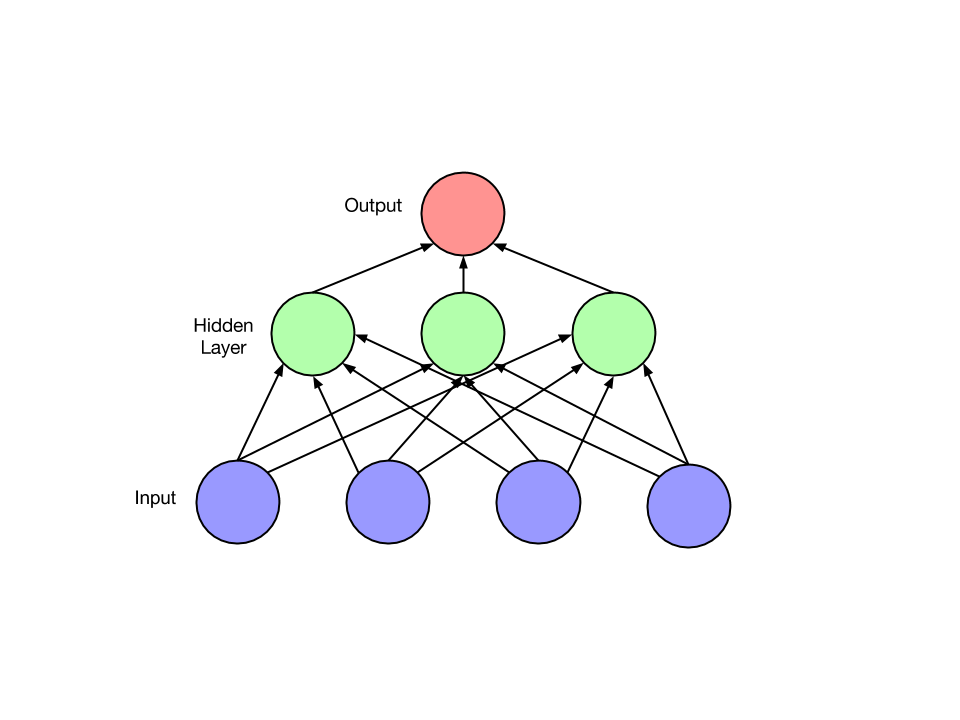}
  \caption{A feedforward neural network. 
  An example is presented to the network by setting the values of the blue (bottom) nodes. 
  The values of the nodes in each layer are computed successively as a function of the prior layers 
  until output is produced at the topmost layer.}
  \label{fig:feedforward-neural-network}
\end{figure}

\subsection{Feedforward networks and backpropagation}

With a neural model of computation, one must determine the order in which computation should proceed.
Should nodes be sampled one at a time and updated, 
or should the value of all nodes be calculated at once and then all updates applied simultaneously?
Feedforward networks (Figure~\ref{fig:feedforward-neural-network}) are a restricted class of networks which 
deal with this problem by forbidding cycles in the directed graph of nodes.
Given the absence of cycles, all nodes can be arranged into layers,
and the outputs in each layer can be calculated given the outputs from the lower layers.

The input $\boldsymbol{x}$ to a feedforward network is provided 
by setting the values of the lowest layer.
Each higher layer is then successively computed 
until output is generated at the topmost layer $\boldsymbol{\hat{y}}$.
Feedforward networks are frequently used for supervised learning tasks 
such as classification and regression.
Learning is accomplished by iteratively updating each of the weights to minimize a loss function, 
$\mathcal{L}(\boldsymbol{\hat{y}},\boldsymbol{y})$, 
which penalizes the distance between the output $\boldsymbol{\hat{y}}$ 
and the target $\boldsymbol{y}$.


The most successful algorithm for training neural networks is backpropagation,
introduced for this purpose by \citet{rumelhart1985learning}.
Backpropagation uses the chain rule to calculate the derivative of the loss function $\mathcal{L}$
with respect to each parameter in the network.
The weights are then adjusted by gradient descent. 
Because the loss surface is non-convex, there is no assurance 
that backpropagation will reach a global minimum.
Moreover, exact optimization is known to be an NP-hard problem.
However, a large body of work on heuristic pre-training and optimization techniques
has led to impressive empirical success on many supervised learning tasks.
In particular, convolutional neural networks, popularized by \citet{le1990handwritten},
are a variant of feedforward neural network that holds records since 2012 
in many computer vision tasks such as object detection \citep{krizhevsky2012imagenet}.

Nowadays, neural networks are usually trained with stochastic gradient descent (SGD) using mini-batches. 
With batch size equal to one,
the stochastic gradient update equation is
$$\boldsymbol{w} \gets \boldsymbol{w} -  \eta \nabla_{\boldsymbol{w}} F_i$$
where $\eta$ is the learning rate 
and $\nabla_{\boldsymbol{w}} F_i$ is the gradient of the objective function 
with respect to the parameters $\boldsymbol{w}$ as calculated on a single example $(x_i, y_i)$.
Many variants of SGD are used to accelerate learning.
Some popular heuristics, such as AdaGrad \citep{duchi2011adaptive}, AdaDelta \citep{zeiler2012adadelta}, 
and RMSprop \citep{rmsprop},  tune the learning rate adaptively for each feature.
AdaGrad, arguably the most popular, 
adapts the learning rate by caching the sum of squared gradients 
with respect to each parameter at each time step. 
The step size for each feature is multiplied by the inverse of the square root of this cached value.
AdaGrad leads to fast convergence on convex error surfaces, 
but because the cached sum is monotonically increasing, the method has a monotonically decreasing learning rate,
which may be undesirable on highly non-convex loss surfaces.
RMSprop modifies AdaGrad by introducing a decay factor in the cache, 
changing the monotonically growing value into a moving average. 
Momentum methods are another common SGD variant used to train neural networks.
These methods add to each update a decaying sum of the previous updates. 
When the momentum parameter is tuned well and the network is initialized well,
momentum methods can train deep nets and recurrent nets 
competitively with more computationally expensive methods like the Hessian-free optimizer of \citet{sutskever2013importance}.

To calculate the gradient in a feedforward neural network, backpropagation proceeds as follows. 
First, an example is  propagated forward through the network to produce a
value $v_j$ at each node and outputs $\boldsymbol{\hat{y}}$ at the topmost layer.
Then, a loss function value $\mathcal{L}(\hat{y}_k, y_k)$ is computed at each output node $k$.
Subsequently, for each output node $k$, we calculate
$$\delta_k = \frac{\partial \mathcal{L}(\hat{y}_k, y_k)}{\partial \hat{y}_k} \cdot l_k'(a_k).$$ 
Given these values $\delta_k$, for each node in the immediately prior layer we calculate
$$\delta_j = l'(a_j) \sum_{k} \delta_k \cdot w_{kj}.$$
This calculation is performed successively for each lower layer to yield
$\delta_j$ for every node $j$ given the $\delta$ values for each node connected to $j$ by an outgoing edge.
Each value $\delta_j$ represents the derivative $\partial \mathcal{L}/{\partial a_j}$ of the total loss function
with respect to that node's {incoming activation}. 
Given the values $v_j$ calculated during the forward pass, 
and the values $\delta_j$ calculated during the backward pass, 
the derivative of the loss $\mathcal{L}$ with respect a given parameter $w_{jj'}$ is
$$\frac{\partial \mathcal{L}}{\partial w_{jj'}}= \delta_j  v_{j'}.$$

Other methods have been explored for learning the weights in a neural network.
A number of papers from the 1990s \citep{belew1990evolving, gruau1994neural}
championed the idea of learning neural networks with genetic algorithms,
with some even claiming that achieving success on real-world problems 
only by applying many small changes to the weights of a network was impossible.
Despite the subsequent success of backpropagation, interest in genetic algorithms continues.
Several recent papers explore genetic algorithms for neural networks, 
especially as a means of learning the architecture of neural networks,
 a problem not addressed by backpropagation \citep{bayer2009evolving, harp2013optimizing}.
By \emph{architecture} we mean the number of layers, 
the number of nodes in each, the connectivity pattern among the layers, the choice of activation functions, etc.

One open question in neural network research is how to exploit sparsity in training.
In a neural network with sigmoidal or $\textit{tanh}$ activation functions,
the nodes in each layer never take value exactly zero.
Thus, even if the inputs are sparse, the nodes at each hidden layer are not.
However, rectified linear units (ReLUs) introduce sparsity to hidden layers \citep{glorot2011deep}.
In this setting, a promising path may be to store the sparsity pattern when computing each layer's values
and use it to speed up computation of the next layer in the network.
Some recent work 
shows that given sparse inputs to a linear model with a standard regularizer, 
sparsity can be fully exploited even if regularization makes the gradient be not sparse
\citep{carpenter2008lazy, langford2009sparse, singer2009efficient, lipton2015efficient}.

\section{Recurrent neural networks}

\begin{figure}[t]
  \centering
  \includegraphics[width=.7\linewidth]{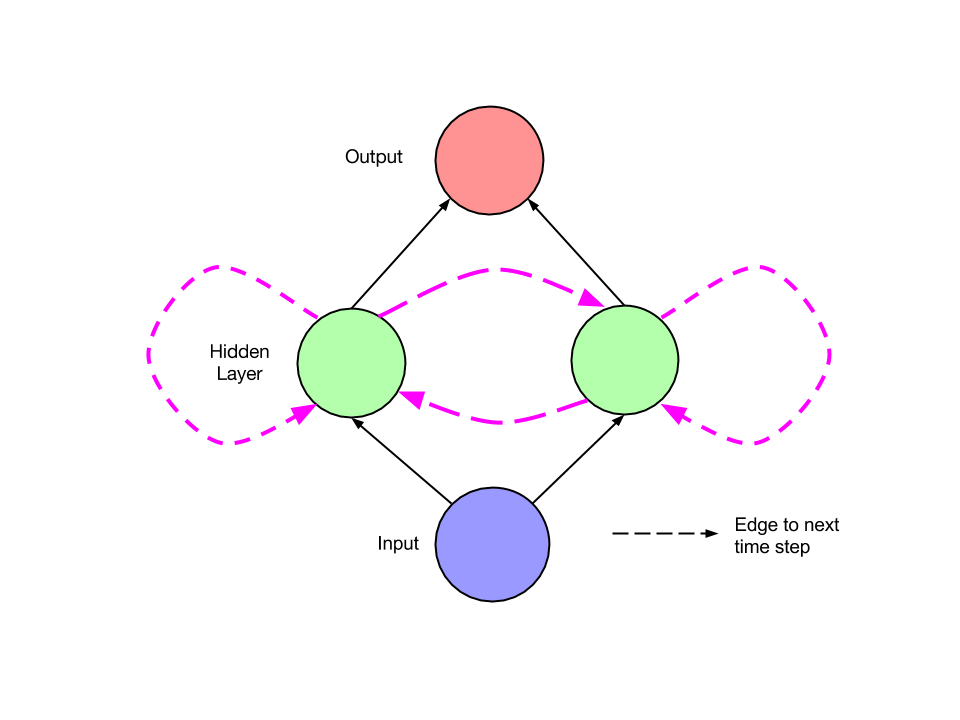}
  \caption{A simple recurrent network.
  At each time step $t$, activation is passed along solid edges as in a feedforward network. 
  Dashed edges connect a source node at each time $t$ to a target node at each following time $t+1$.}
  \label{fig:simple-recurrent}
\end{figure}

Recurrent neural networks are 
feedforward neural networks augmented by the inclusion of edges that span adjacent time steps,
introducing a notion of time to the model. 
Like feedforward networks, RNNs may not have cycles among conventional edges.
However, edges that connect adjacent time steps,
called recurrent edges, may form cycles, including cycles of length one
that are self-connections from a node to itself across time.
At time $t$, nodes with recurrent edges 
receive {input} from the current data point $\boldsymbol{x}^{(t)}$
and also from hidden node values $\boldsymbol{h}^{(t-1)}$ in the network's previous state. 
The output $\boldsymbol{\hat{y}}^{(t)}$ at each time $t$ is calculated 
given the hidden node values $\boldsymbol{h}^{(t)}$ at time $t$. 
Input $\boldsymbol{x}^{(t-1)}$ at time $t-1$
can influence the output $\boldsymbol{\hat{y}}^{(t)}$ at time $t$ and later
by way of the recurrent connections.

Two equations specify all calculations necessary for computation at each time step on the forward pass
in a simple recurrent neural network as in Figure~\ref{fig:simple-recurrent}:
$$
\boldsymbol{h}^{(t)} = \sigma(W^{\mbox{hx}} \boldsymbol{x}^{(t)} + W^{\mbox{hh}} \boldsymbol{h}^{(t-1)} +\boldsymbol{b}_h  ) 
$$
$$
\hat{\boldsymbol{y}}^{(t)} = \softmax ( W^{\mbox{yh}}  \boldsymbol{h}^{(t)} + \boldsymbol{b}_y).
$$
Here $W_{\mbox{hx}}$ is the matrix of conventional weights between the input and the hidden layer
and $W_{\mbox{hh}}$ is the matrix of recurrent weights between the hidden layer and itself at adjacent time steps.
The vectors $\boldsymbol{b}_h$ and $\boldsymbol{b}_y$ are bias parameters which allow each node to learn an offset.

\begin{figure}[t]
  \centering
  \includegraphics[width=.7\linewidth]{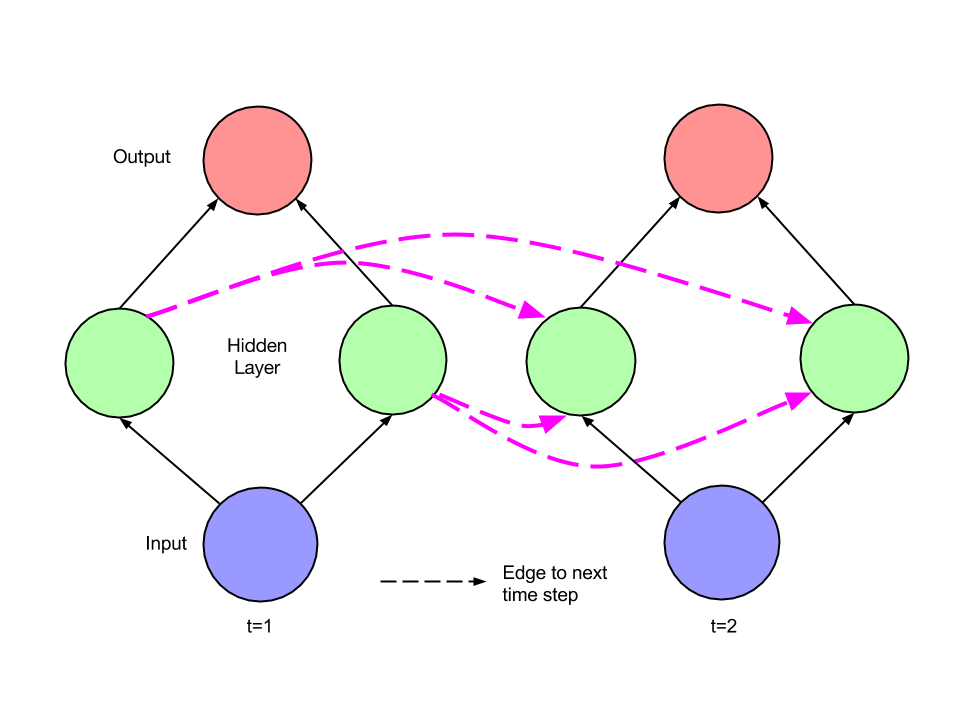}
  \caption{The recurrent network of Figure~\ref{fig:simple-recurrent} unfolded across time steps.}
  \label{fig:unfolded-rnn}
\end{figure}


The dynamics of the network depicted in Figure~\ref{fig:simple-recurrent} across time steps can be visualized 
by \emph{unfolding} it as in Figure~\ref{fig:unfolded-rnn}. 
Given this picture, the network can be interpreted not as cyclic, but rather as a deep network
with one layer per time step and shared weights across time steps.
It is then clear that the unfolded network can be trained across many time steps using backpropagation.
This algorithm, called \emph{backpropagation through time} (BPTT),
was introduced by \citet{werbos1990backpropagation}.
All recurrent networks in common current use apply it.

\begin{figure}[t]
  \centering
  \includegraphics[width=.7\linewidth]{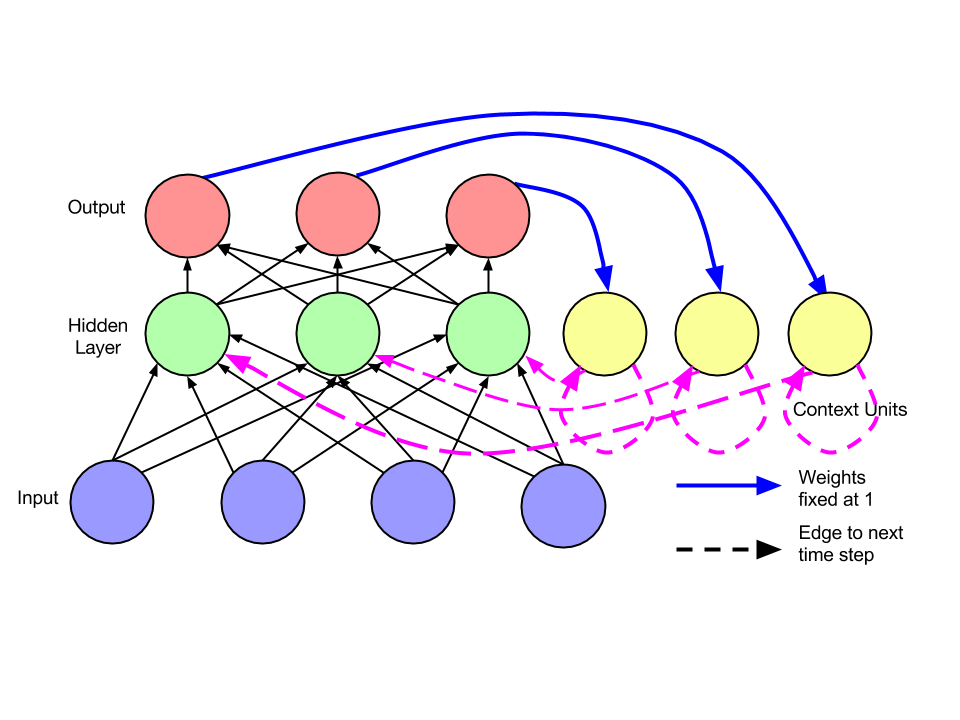}
  \caption{A recurrent neural network as proposed by \citet{jordan1997serial}.
  Output units are connected to special units that at the next time step
  feed into themselves and into hidden units.}
  \label{fig:jordan-network}
\end{figure}

\begin{figure}[t]
  \centering
  \includegraphics[width=.7\linewidth]{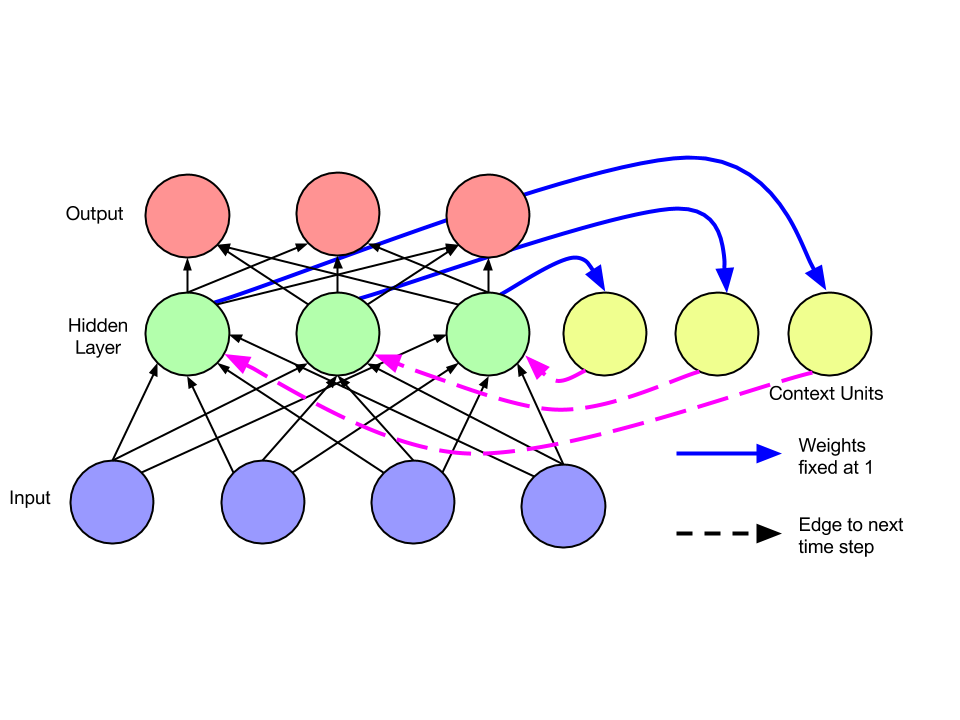}
  \caption{A recurrent neural network as described by \citet{elman1990finding}. 
  Hidden units are connected to context units, 
  which feed back into the hidden units at the next time step.}
  \label{fig:elman-network}
\end{figure}

\subsection{Early recurrent network designs}

The foundational research on recurrent networks took place in the 1980s. 
In 1982, Hopfield introduced a family of recurrent neural networks 
that have pattern recognition capabilities \citep{hopfield1982neural}.
They are defined by the values of the weights between nodes
and the link functions are simple thresholding at zero.
In these nets, a pattern is placed in the network by setting the values of the nodes.
The network then runs for some time according to its update rules, 
and eventually another pattern is read out.
Hopfield networks are useful for recovering a stored pattern from a corrupted version
and are the forerunners of Boltzmann machines and auto-encoders.

An early architecture for supervised learning on sequences was introduced by \citet{jordan1997serial}. 
Such a network (Figure~\ref{fig:jordan-network}) is a feedforward network with a single hidden layer
that is extended with {special units}.%
\footnote{
\citet{jordan1997serial} calls the special units ``state units"
while \citet{elman1990finding} calls a corresponding structure ``context units." 
In this paper we simplify terminology by using only ``context units".
}
Output node values are fed to the special units, 
which then feed these values to the hidden nodes at the following time step.
If the output values are actions, the special units allow the network 
to {remember} actions taken at previous time steps.
Several modern architectures use a related form of direct transfer from output nodes;
\citet{sutskever2014sequence} translates sentences between natural languages,
and when generating a text sequence, the word chosen at each time step 
is fed into the network as input at the following time step.
Additionally, the special units in a Jordan network are self-connected.
Intuitively, these edges allow sending information across multiple time steps 
without perturbing the output at each intermediate time step.

The architecture introduced by \citet{elman1990finding}
is simpler than the earlier Jordan architecture. 
Associated with each unit in the hidden layer is a context unit.
Each such unit $j'$ takes as input the state 
of the corresponding hidden node $j$ at the previous time step,
along an edge of fixed weight $w_{j'j} = 1$.
This value then feeds back into the same hidden node $j$ along a standard edge.
This architecture is equivalent to a simple RNN in which each hidden node 
has a single self-connected recurrent edge.
The idea of fixed-weight recurrent edges that make hidden nodes self-connected
is fundamental in subsequent work on LSTM networks \citep{hochreiter1997long}.

\citet{elman1990finding} trains the network using backpropagation 
and demonstrates that the network can learn time dependencies. 
The paper features two sets of experiments. 
The first extends the logical operation \emph{exclusive or} (XOR) 
to the time domain by concatenating sequences of three tokens.
For each three-token segment, e.g.~``011", the first two tokens (``01") are chosen randomly 
and the third (``1") is set by performing xor on the first two.
Random guessing should achieve accuracy of $50\%$.
A perfect system should perform the same as random for the first two tokens,
but guess the third token perfectly, achieving accuracy of $66.7\%$.
The simple network of \citet{elman1990finding} does in fact approach this maximum achievable score.

\begin{figure}
  \center
  \includegraphics[width=.7\linewidth]{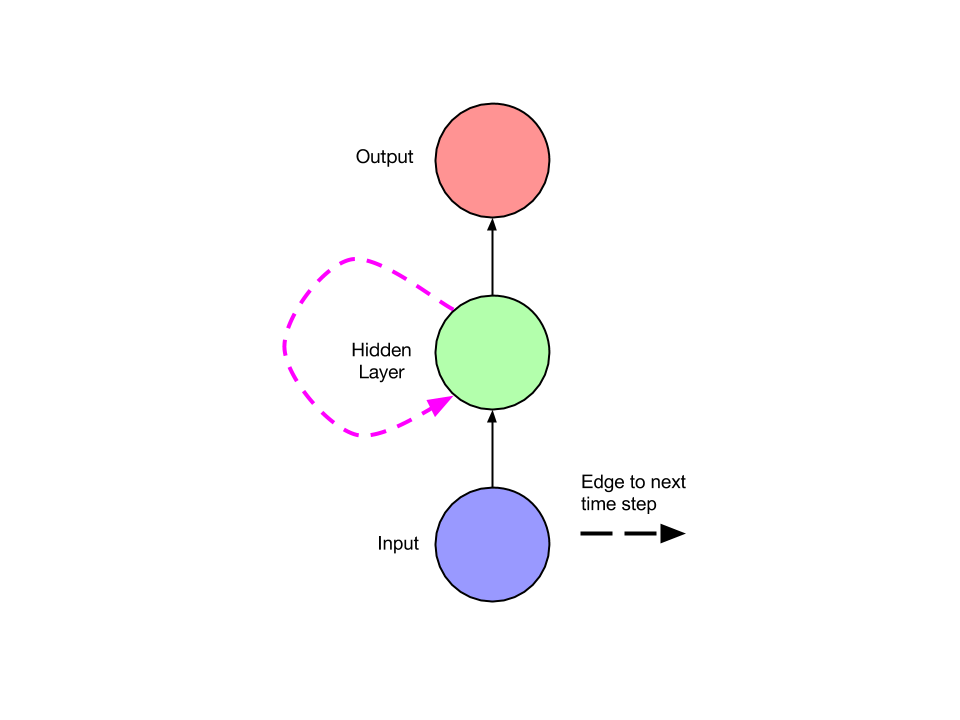}
  \caption{A simple recurrent net with one input unit, one output unit, and one recurrent hidden unit.}
  \label{fig:vanishing1}
\end{figure}

\subsection{Training recurrent networks}

Learning with recurrent networks has long been considered to be difficult. 
Even for standard feedforward networks, the optimization task is NP-complete \cite{blum1993training}. 
But learning with recurrent networks can be especially challenging 
due to the difficulty of learning long-range dependencies, as
described by \citet{bengio1994learning} and 
expanded upon by \citet{hochreiter2001gradient}.
The problems of \emph{vanishing} and \emph{exploding} gradients
occur when backpropagating errors across many time steps.
As a toy example, consider a network with a single input node, 
a single output node, and a single recurrent hidden node (Figure~\ref{fig:vanishing1}).
Now consider an input passed to the network at time $\tau$ and an error calculated at time $t$,
assuming input of zero in the intervening time steps.
The tying of weights across time steps means that the recurrent edge at the hidden node $j$ always has the same weight.
Therefore, the contribution of the input at time $\tau$ to the output at time $t$ 
will either explode or approach zero, exponentially fast, as $t - \tau$ grows large.
Hence the derivative of the error with respect to the input 
will either explode or vanish.

\begin{figure}
  \center
  \includegraphics[width=.7\linewidth]{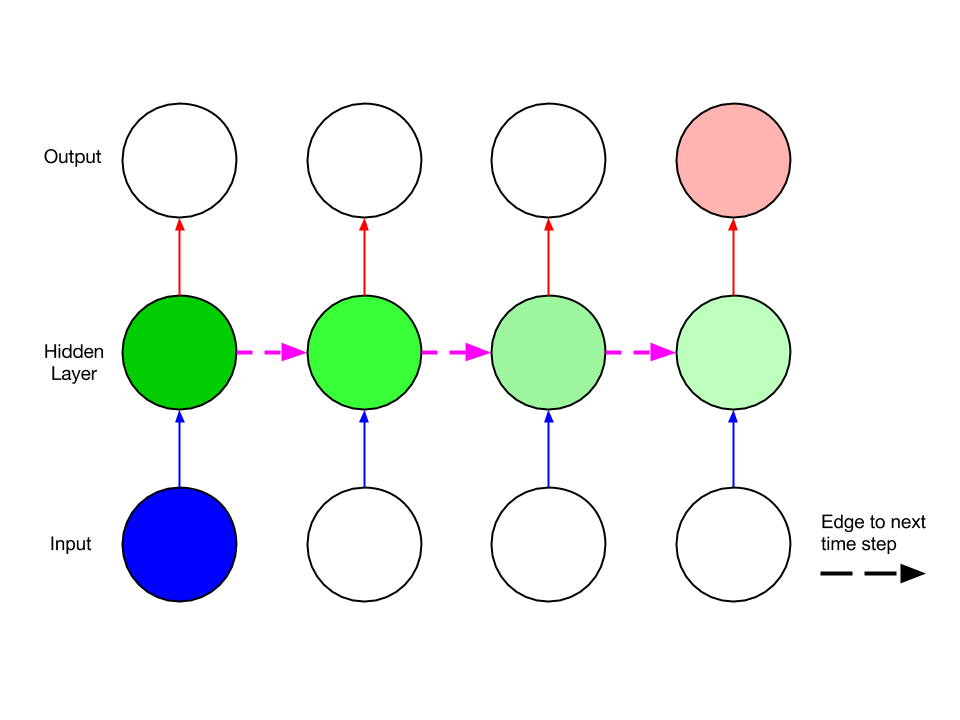}
  \caption{A visualization of the vanishing gradient problem, 
using the network depicted in Figure~\ref{fig:vanishing1}, adapted from \citet{graves2012supervised}. 
If the weight along the recurrent edge is less than one, 
the contribution of the input at the first time step 
to the output at the final time step 
will decrease exponentially fast as a function of the length of the time interval in between.
}
  \label{fig:vanishing2}
\end{figure}

Which of the two phenomena occurs
depends on whether the weight of the recurrent edge $|w_{jj}| > 1$ or $|w_{jj}| < 1$ 
and on the activation function in the hidden node (Figure~\ref{fig:vanishing2}). 
Given a sigmoid activation function, 
the vanishing gradient problem is more pressing, 
but with a rectified linear unit $\max(0,x)$, 
it is easier to imagine the exploding gradient.
\citet{pascanu2012difficulty} give a thorough mathematical treatment 
of the vanishing and exploding gradient problems,
characterizing exact conditions under which these problems may occur.
Given these conditions, they suggest an approach to training via a regularization term
that forces the weights to values where the gradient neither vanishes nor explodes.

Truncated backpropagation through time (TBPTT) is one solution to the exploding gradient problem 
for continuously running networks \citep{williams1989learning}.
With TBPTT, some maximum number of time steps is set along which error can be propagated.
While TBPTT with a small cutoff can be used to alleviate the exploding gradient problem, 
it requires that one sacrifice the ability to learn long-range dependencies.
The LSTM architecture described below 
uses carefully designed nodes with recurrent edges with fixed unit weight
as a solution to the vanishing gradient problem.

The issue of local optima is an obstacle to effective training
that cannot be dealt with simply by modifying the network architecture.
Optimizing even a single hidden-layer feedforward network is 
an NP-complete problem \citep{blum1993training}.
However, recent empirical and theoretical studies suggest that 
in practice, the issue may not be as important as once thought. 
\citet{dauphin2014identifying} show that while many critical points exist 
on the error surfaces of large neural networks, 
the ratio of saddle points to true local minima increases exponentially with the size of the network,
and algorithms can be designed to escape from saddle points.

Overall, along with the improved architectures explained below,
fast implementations and better gradient-following heuristics 
have rendered RNN training feasible.
Implementations of forward and backward propagation using GPUs, 
such as the Theano \citep{bergstra2010theano} and Torch \citep{collobert2011torch7} packages,
have made it straightforward to implement fast training algorithms.
In 1996, prior to the introduction of the LSTM, attempts to train recurrent nets to bridge long time gaps 
were shown to perform no better than random guessing \citep{hochreiter1996bridging}.
However, RNNs are now frequently trained successfully.

For some tasks, 
freely available software can be run on a single GPU 
and produce compelling results in hours \citep{karpathyunreasonable}.
\citet{martens2011learning} reported success training recurrent neural networks 
with a Hessian-free truncated Newton approach,
and applied the method to a network which learns to generate text one character at a time in \citep{sutskever2011generating}.
In the paper that describes the abundance of saddle points 
on the error surfaces of neural networks \citep{dauphin2014identifying}, 
the authors present a saddle-free version of Newton's method.
Unlike Newton's method, which is attracted to critical points, including saddle points,
this variant is specially designed to escape from them.
Experimental results include a demonstration 
of improved performance on recurrent networks.
Newton's method requires computing the Hessian, 
which is prohibitively expensive for large networks, 
scaling quadratically with the number of parameters.
While their algorithm only approximates the Hessian, 
it is still computationally expensive compared to SGD.
Thus the authors describe a hybrid approach 
in which the saddle-free Newton method is applied only
in places where SGD appears to be {stuck}.

\section{Modern RNN architectures}

The most successful RNN architectures for sequence learning stem from two papers published in 1997.
The first paper, \emph{Long Short-Term Memory} by \citet{hochreiter1997long},
introduces the memory cell, a unit of computation that replaces 
traditional nodes in the hidden layer of a network.
With these memory cells, networks are able to overcome difficulties with training 
encountered by earlier recurrent networks.
The second paper, \emph{Bidirectional Recurrent Neural Networks} by \citet{schuster1997bidirectional},
introduces an architecture in which information from both the future and the past 
are used to determine the output at any point in the sequence.
This is in contrast to previous networks, in which only past input can affect the output,
and has been used successfully for sequence labeling tasks in natural language processing, among others.
Fortunately, the two innovations are not mutually exclusive, 
and have been successfully combined for phoneme classification \citep{graves2005framewise} 
and handwriting recognition \citep{graves2009novel}.
In this section we explain the LSTM and BRNN
and we describe the \emph{neural Turing machine} (NTM), 
which extends RNNs with an addressable external memory \citep{graves2014neural}.

\subsection{Long short-term memory  (LSTM)}

\begin{figure}[t]
  \center
  \includegraphics[width=.8\linewidth]{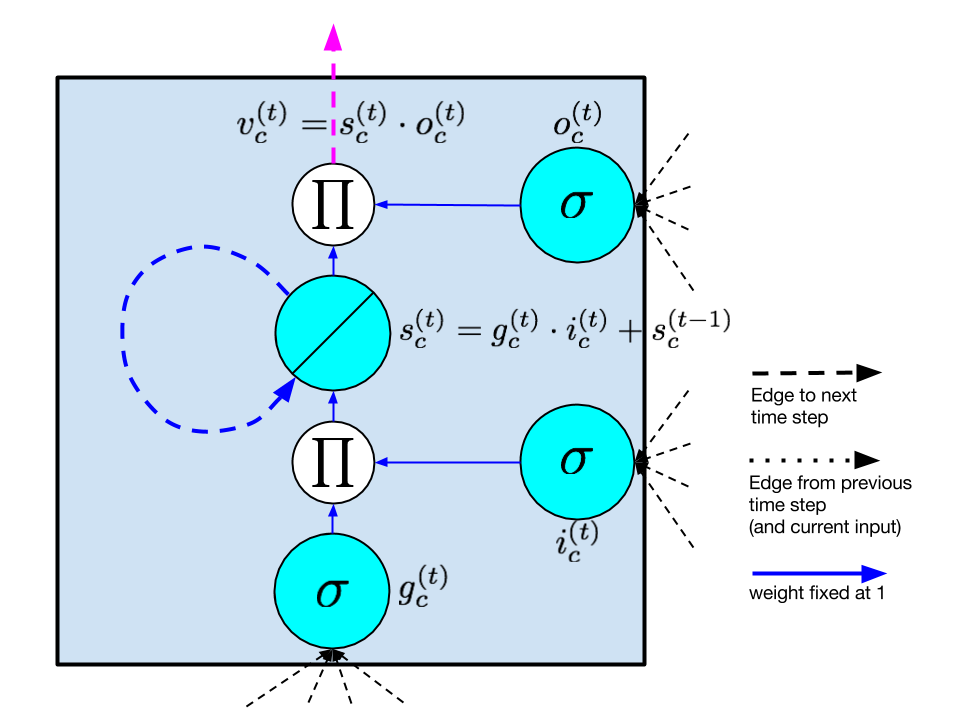}
  \caption{One LSTM memory cell as proposed by \citet{hochreiter1997long}. 
The self-connected node is the internal state $s$. 
The diagonal line indicates that it is linear, i.e.~the identity link function is applied. 
The blue dashed line is the recurrent edge, which has fixed unit weight.
Nodes marked $\Pi$ output the product of their inputs. 
All edges into and from $\Pi$ nodes also have fixed unit weight.}
\label{fig:lstm}
\end{figure}

\citet{hochreiter1997long} introduced the LSTM model 
primarily in order to overcome the problem of vanishing gradients.
This model resembles a standard recurrent neural network with a hidden layer,
but each ordinary node (Figure~\ref{fig:artificial-neuron}) in the hidden layer
is replaced by a \emph{memory cell} (Figure~\ref{fig:lstm}). 
Each memory cell contains a node with a self-connected recurrent edge of fixed weight one,
ensuring that the gradient can pass across many time steps without vanishing or exploding.
To distinguish references to a memory cell 
and not an ordinary node, we use the subscript $c$.

The term ``long short-term memory" comes from the following intuition.
Simple recurrent neural networks have \emph{long-term memory} in the form of weights.
The weights change slowly during training, encoding general knowledge about the data.
They also have \emph{short-term memory} in the form of ephemeral activations,
which pass from each node to successive nodes.
The LSTM model introduces an intermediate type of storage via the memory cell.
A memory cell is a composite unit, built from simpler nodes in a specific connectivity pattern,
with the novel inclusion of multiplicative nodes, represented in diagrams by the letter $\Pi$. 
All elements of the LSTM cell are enumerated and described below.
%
Note that when we use vector notation, 
we are referring to the values of the nodes in an entire layer of cells.
For example, $\boldsymbol{s}$ is a vector containing the value of $s_c$ at each memory cell $c$ in a layer.
When the subscript $c$ is used, it is to index an individual memory cell.

\begin{itemize}
\item \emph{Input node:}
This unit, labeled $g_c$, is a node that takes activation in the standard way
from the input layer $\boldsymbol{x^{(t)}}$ at the current time step
and (along recurrent edges) from the hidden layer at the previous time step $\boldsymbol{h}^{(t-1)}$.
Typically, the summed weighted input is run through a $\textit{tanh}$ activation function,
although in the original LSTM paper, the activation function is a $\textit{sigmoid}$.

\item \emph{Input gate:}
Gates are a distinctive feature of the LSTM approach.
A gate is a sigmoidal unit that, like the {input node}, takes 
activation from the current data point $\boldsymbol{x}^{(t)}$ 
as well as from the hidden layer at the previous time step. 
A gate is so-called because its value is used to multiply the value of another node. 
It is a \emph{gate} in the sense that if its value is zero, then flow from the other node is cut off.
If the value of the gate is one, all flow is passed through.
The value of the \emph{input gate} $i_c$ multiplies the value of the $\emph{input node}$.

\item \emph{Internal state:}
At the heart of each memory cell is a node $s_c$ with linear activation, 
which is referred to in the original paper as the ``internal state" of the cell.
The internal state $s_c$ has a self-connected recurrent edge with fixed unit weight. 
Because this edge spans adjacent time steps with constant weight,
error can flow across time steps without vanishing or exploding.
This edge is often called the \emph{constant error carousel}.
In vector notation, the update for the internal state is
$\boldsymbol{s}^{(t)} = \boldsymbol{g}^{(t)} \odot \boldsymbol{i}^{(t)} + \boldsymbol{s}^{(t-1)}$
where $\odot$ is pointwise multiplication.

\item \emph{Forget gate:}
{These gates} $f_c$ were introduced by \citet{gers2000learning}.
They provide a method by which the network can learn to flush the contents of the internal state.
This is especially useful in continuously running networks.
With forget gates, the equation to calculate the internal state on the forward pass is
$$
\boldsymbol{s}^{(t)} = \boldsymbol{g}^{(t)} \odot \boldsymbol{i}^{(t)} 
+ \boldsymbol{f}^{(t)} \odot \boldsymbol{s}^{(t-1)}.
$$

\item \emph{Output gate:}
The value $v_c$ ultimately produced by a memory cell 
is the value of the internal state $s_c$ multiplied by the value of the \emph{output gate} $o_c$. 
It is customary that the internal state first be run through a \textit{tanh} activation function, 
as this gives the output of each cell the same dynamic range as an ordinary \textit{tanh} hidden unit.
However, in other neural network research, rectified linear units,
which have a greater dynamic range, are easier to train.
Thus it seems plausible that the nonlinear function on the internal state might be omitted.
\end{itemize}

In the original paper and in most subsequent work, the input node is labeled $g$.
We adhere to this convention but note that it may be confusing as $g$ does not stand for \emph{gate}.
In the original paper, the gates are called $y_{in}$ and $y_{out}$ but this is confusing 
because $y$ generally stands for output in the machine learning literature. 
Seeking comprehensibility, we break with this convention and use $i$, $f$, and $o$ to refer to 
input, forget and output gates respectively, as in \citet{sutskever2014sequence}. 

Since the original LSTM was introduced, several variations have been proposed. 
{Forget gates}, described above, were proposed in 2000 and were not part of the original LSTM design.
However, they have proven effective and are standard in most modern implementations.
That same year, \citet{gers2000recurrent} proposed peephole connections
that pass from the internal state directly to the input and output gates of that same node 
without first having to be modulated by the output gate. 
They report that these connections improve performance 
on timing tasks where the network must learn 
to measure precise intervals between events.
The intuition of the peephole connection can be captured by the following example.
Consider a network which must learn to count objects and emit some desired output 
when $n$ objects have been seen.
The network might learn to let some fixed amount of activation into the internal state after each object is seen. 
This activation is trapped in the internal state $s_c$ by the constant error carousel, 
and is incremented iteratively each time another object is seen.
When the $n$th object is seen,
the network needs to know to let out content from the internal state so that it can affect the output.
To accomplish this, the output gate $o_c$ must know the content of the internal state $s_c$.
Thus $s_c$ should be an input to $o_c$.

\begin{figure}
  \center
  \includegraphics[width=.8\linewidth]{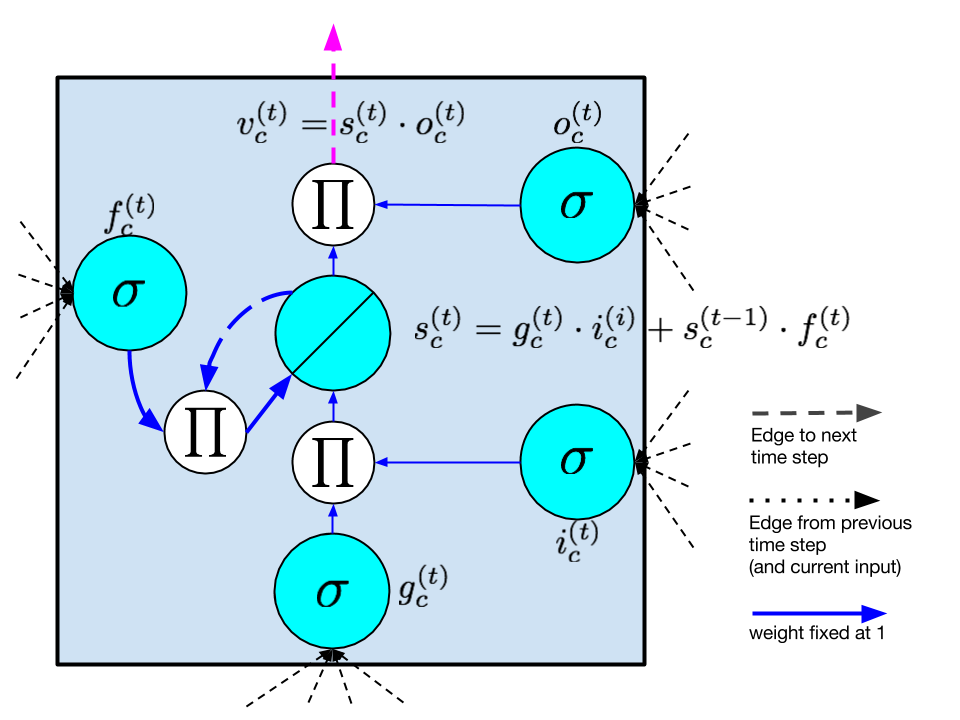}
  \caption{LSTM memory cell with a forget gate as described by \citet{gers2000learning}.}
  \label{fig:lstm-forget}
\end{figure}

Put formally, computation in the LSTM model proceeds according to the following calculations,
which are performed at each time step. 
These equations give the full algorithm for a modern LSTM with forget gates:
$$ \boldsymbol{g}^{(t)} = \phi( W^{\mbox{gx}} \boldsymbol{x}^{(t)} +   W^{\mbox{gh}} \boldsymbol{h}^{(t-1)}  + \boldsymbol{b}_g)$$
$$ \boldsymbol{i}^{(t)}  =    \sigma( W^{\mbox{ix}} \boldsymbol{x}^{(t)} + W^{\mbox{ih}} \boldsymbol{h}^{(t-1)} + \boldsymbol{b}_i) $$
$$ \boldsymbol{f}^{(t)}  =    \sigma( W^{\mbox{fx}} \boldsymbol{x}^{(t)} + W^{\mbox{fh}} \boldsymbol{h}^{(t-1)}  + \boldsymbol{b}_f) $$
$$ \boldsymbol{o}^{(t)} =    \sigma( W^{\mbox{ox}} \boldsymbol{x}^{(t)} + W^{\mbox{oh}} \boldsymbol{h}^{(t-1)}  + \boldsymbol{b}_o) $$
$$ \boldsymbol{s}^{(t)} = \boldsymbol{g}^{(t)} \odot \boldsymbol{i}^{(i)} + \boldsymbol{s}^{(t-1)} \odot \boldsymbol{f}^{(t)}$$
$$ \boldsymbol{h}^{(t)}  = \phi ( \boldsymbol{s}^{(t)}) \odot \boldsymbol{o}^{(t)}. $$ 
The value of the hidden layer of the LSTM at time $t$ is the vector $\boldsymbol{h}^{(t)}$,
while $\boldsymbol{h}^{(t-1)}$ is the values output by each memory cell in the hidden layer at the previous time.
Note that these equations include the forget gate, but not peephole connections.
The calculations for the simpler LSTM without forget gates 
are obtained by setting $\boldsymbol{f}^{(t)} = 1$ for all $t$.  
We use the $\textit{tanh}$ function $\phi$ for the input node $g$
following the state-of-the-art design of \citet{zaremba2014learning}.
However, in the original LSTM paper, the activation function for $g$ is the sigmoid $\sigma$.

\begin{figure}[t]
  \center
  \includegraphics[width=.8\linewidth]{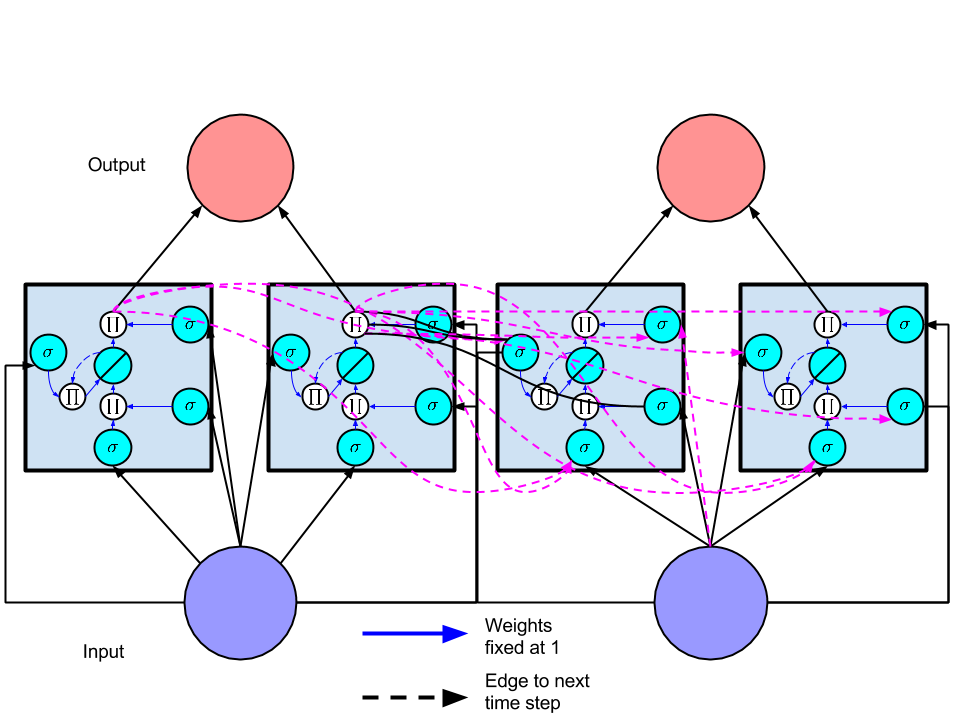}
  \caption{A recurrent neural network with a hidden layer consisting of two memory cells. 
The network is shown unfolded across two time steps.}
  \label{fig:lstm-network}
\end{figure}

Intuitively, in terms of the forward pass, the LSTM can learn when to let activation into the internal state. 
As long as the input gate takes value zero, no activation can get in. 
Similarly, the output gate learns when to let the value out.
When both gates are \emph{closed}, the activation is trapped in the memory cell,
neither growing nor shrinking, nor affecting the output at intermediate time steps.
In terms of the backwards pass, 
the constant error carousel enables the gradient to propagate back across many time steps, 
neither exploding nor vanishing.
In this sense, the gates are learning when to let {error} in, and when to let it out.
In practice, the LSTM has shown a superior ability 
to learn long-range dependencies as compared to simple RNNs.
Consequently, the majority of state-of-the-art application papers covered in this review use the LSTM model.

One frequent point of confusion is the manner in which multiple memory cells 
are used together to comprise the hidden layer of a working neural network.
To alleviate this confusion, we depict in Figure~\ref{fig:lstm-network}
a simple network with two memory cells, analogous to Figure~\ref{fig:unfolded-rnn}.
The output from each memory cell flows in the subsequent time step
to the input node and all gates of each memory cell.
It is common to include multiple layers of memory cells \citep{sutskever2014sequence}.
Typically, in these architectures each layer takes input from the layer below at the same time step
and from the same layer in the previous time step.

\begin{figure}[]
  \center
  \includegraphics[width=.8\linewidth]{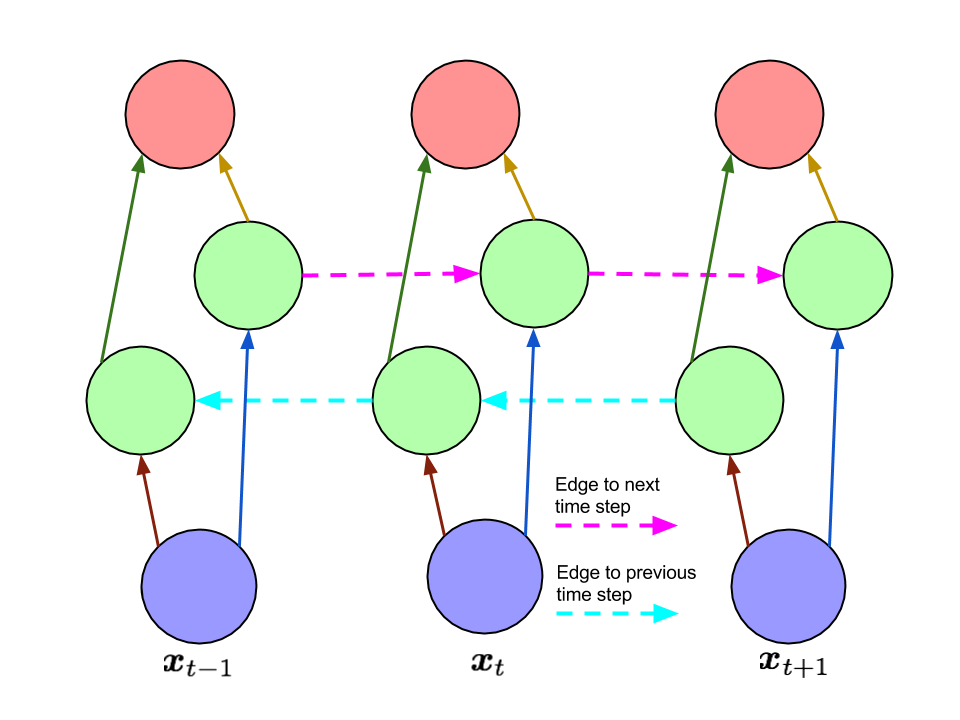}
  \caption{A bidirectional recurrent neural network as described by \citet{schuster1997bidirectional}, 
  unfolded in time.}
  \label{fig:brnn}
\end{figure}

\subsection{Bidirectional recurrent neural networks (BRNNs)}

Along with the LSTM, one of the most used RNN architectures 
is the bidirectional recurrent neural network (BRNN) (Figure~\ref{fig:brnn}) 
first described by \citet{schuster1997bidirectional}. 
In this architecture, there are two layers of hidden nodes.
Both hidden layers are connected to input and output.
The two hidden layers are differentiated in that the first has recurrent connections 
from the past time steps
while in the second the direction of recurrent of connections is flipped,
passing activation backwards along the sequence.
Given an input sequence and a target sequence, 
the BRNN can be trained by ordinary backpropagation after unfolding across time.
The following three equations describe a BRNN:
$$\boldsymbol{h}^{(t)} = \sigma(W^{\mbox{h}\mbox{x}} \boldsymbol{x}^{(t)} + W^{\mbox{h}\mbox{h}} \boldsymbol{h}^{(t-1)} +\boldsymbol{b}_{h}  ) $$
$$\boldsymbol{z}^{(t)} = \sigma(W^{\mbox{z}\mbox{x}} \boldsymbol{x}^{(t)} + W^{\mbox{z}\mbox{z}} \boldsymbol{z}^{(t+1)} +\boldsymbol{b}_{z}  ) $$
$$ \hat{\boldsymbol{y}}^{(t)} = \softmax ( W^{\mbox{yh}} \boldsymbol{h}^{(t)} + W^{\mbox{yz}} \boldsymbol{z}^{(t)} + \boldsymbol{b}_y)$$
where $\boldsymbol{h}^{(t)}$ and $\boldsymbol{z}^{(t)}$ 
are the values of the hidden layers in the forwards and backwards directions respectively.

One limitation of the BRNN is that cannot run continuously, as it requires a fixed endpoint in both the future and in the past.
Further, it is not an appropriate machine learning algorithm for the online setting,
as it is implausible to receive information from the future, i.e., to know sequence elements that have not been observed. 
But for prediction over a sequence of fixed length, it is often sensible to take into account both past and future sequence elements. 
Consider the natural language task of {part-of-speech tagging}. 
Given any word in a sentence, information about both 
the words which precede and those which follow it 
is useful for predicting that word's part-of-speech.

The LSTM and BRNN are in fact compatible ideas. 
The former introduces a new basic unit from which to compose a hidden layer, 
while the latter concerns the wiring of the hidden layers, regardless of what nodes they contain.
Such an approach, termed a BLSTM has been used to achieve state of the art results 
on handwriting recognition and phoneme classification \citep{graves2005framewise, graves2009novel}.

\subsection{Neural Turing machines}

The neural Turing machine (NTM) extends recurrent neural networks 
with an addressable external memory \citep{graves2014neural}. 
This work improves upon the ability of RNNs
to perform complex algorithmic tasks such as sorting.
The authors take inspiration from theories in cognitive science, 
which suggest humans possess a ``central executive" that interacts with a memory buffer \citep{baddeley1996working}. 
By analogy to a Turing machine, in which a program directs \emph{read heads} and \emph{write heads} 
to interact with external memory in the form of a tape, the model is named a Neural Turing Machine.
While technical details of the read/write heads are beyond the scope of this review, 
we aim to convey a high-level sense of the model and its applications.

The two primary components of an NTM are a \emph{controller} and \emph{memory matrix}.
The controller, which may be a recurrent or feedforward neural network, 
takes input and returns output to the outside world,
as well as passing instructions to and reading from the memory.
The memory is represented by a large matrix of $N$ memory locations, 
each of which is a vector of dimension $M$.
Additionally, a number of read and write heads facilitate the interaction 
between the {controller} and the {memory matrix}.
Despite these additional capabilities, the NTM is differentiable end-to-end
and can be trained by variants of stochastic gradient descent using BPTT. 

\citet{graves2014neural} select five algorithmic tasks
 to test the performance of the NTM model. 
By \emph{algorithmic} we mean that for each task, 
the target output for a given input 
can be calculated by following a simple program,
as might be easily implemented in any universal programming language.  
One example is the \emph{copy} task,
where the input is a sequence of fixed length binary vectors followed by a delimiter symbol.
The target output is a copy of the input sequence.  
In another task, \emph{priority sort}, an input consists of a sequence of binary vectors 
together with a distinct scalar priority value for each vector.  
The target output is the sequence of vectors sorted by priority. 
The experiments test whether an NTM can be trained via supervised learning
to implement these common algorithms correctly and efficiently. 
Interestingly, solutions found in this way generalize reasonably well 
to inputs longer than those presented in the training set.  
In contrast, the LSTM without external memory
does not generalize well to longer inputs.
The authors compare three different architectures, namely
an LSTM RNN, an NTM with a feedforward controller, 
and an NTM with an LSTM controller.  
On each task, both NTM architectures significantly outperform
the LSTM RNN both in training set performance and in generalization to test data.

\section{Applications of LSTMs and BRNNs}

\begin{figure}
  \centering
  \includegraphics[width=.6\linewidth]{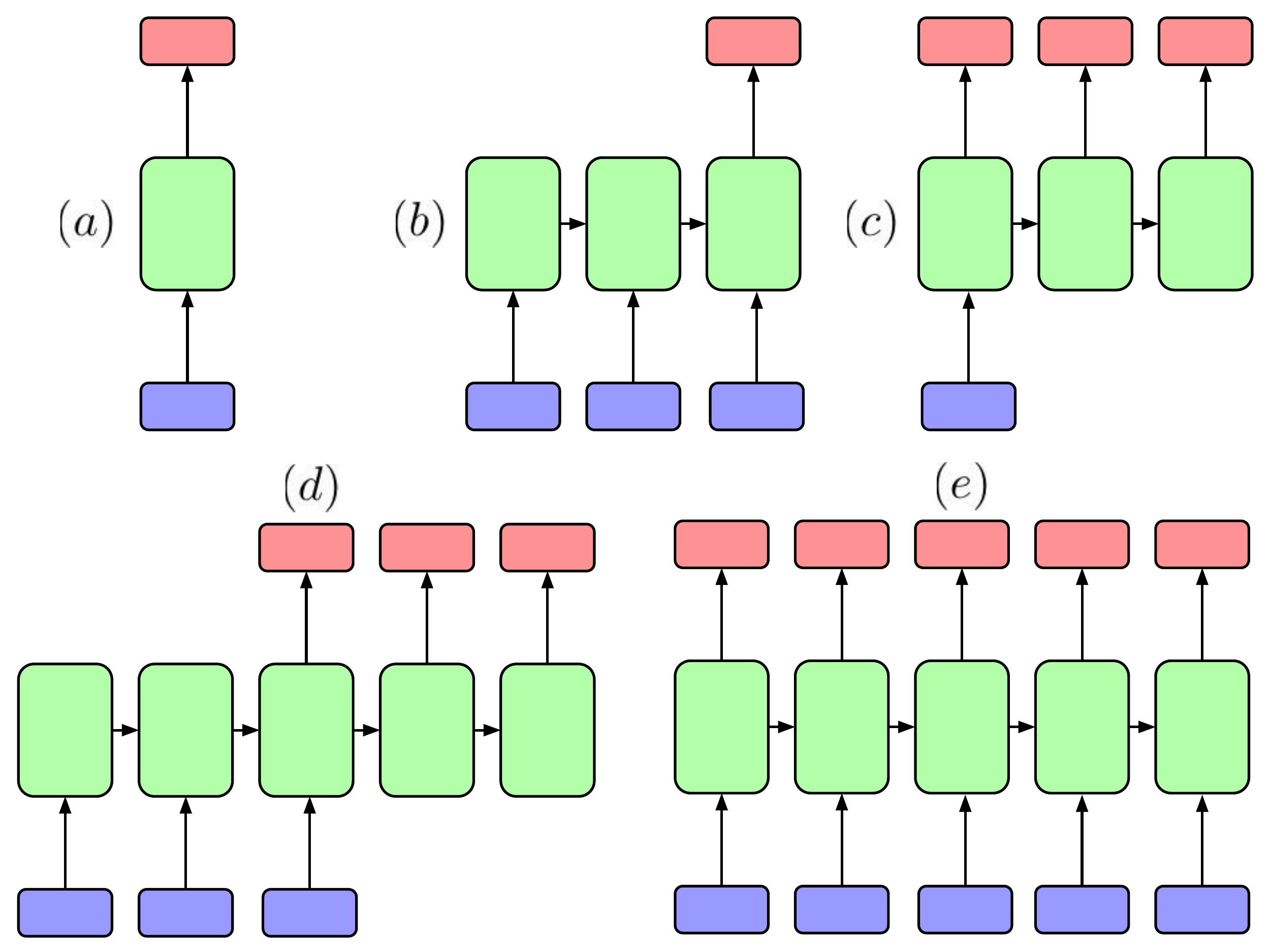}
  \caption{Recurrent neural networks have been used successfully to model  both sequential inputs and sequential outputs 
  as well as mappings between single data points and sequences (in both directions). 
 This figure, based on a similar figure in \citet{karpathyunreasonable}  
 shows how numerous tasks can be modeled with RNNs with sequential inputs and/or sequential outputs.
 In each subfigure, blue rectangles correspond to inputs, 
red rectangles to outputs and green rectangles to the entire hidden state of the neural network.
(a) This is the conventional independent case, as assumed by standard feedforward networks. 
(b) Text and video classification are tasks in which a sequence is mapped to one fixed length vector. 
(c) Image captioning presents the converse case,
where the input image is a single non-sequential data point.
(d) This architecture has been used for natural language translation, 
 a sequence-to-sequence task in which the two sequences may have varying and different lengths. 
(e) This architecture has been used to learn a generative model for text, predicting at each step the following character.}
  \label{fig:rnn-types}
\end{figure}

The previous sections introduced the building blocks 
from which nearly all state-of-the-art recurrent neural networks are composed.
This section looks at several application areas where recurrent networks 
have been employed successfully.
Before describing state of the art results in detail,
it is appropriate to convey a concrete sense of the precise architectures with which 
many important tasks can be expressed clearly as sequence learning problems with recurrent neural networks.
Figure \ref{fig:rnn-types} demonstrates several common RNN architectures 
and associates each with corresponding well-documented tasks.

In the following subsections, we first introduce the representations of natural language
used for input and output to recurrent neural networks
and the commonly used performance metrics for sequence prediction tasks.
Then we survey state-of-the-art results in machine translation, 
image captioning, video captioning, and handwriting recognition. 
Many applications of RNNs involve  processing written language.
Some applications, such as image captioning, 
involve generating strings of text.
Others, such as machine translation and dialogue systems, 
require both inputting and outputting text.
Systems which output text are more difficult to evaluate empirically than those which
produce binary predictions or numerical output.
As a result several methods have been developed to assess the quality of translations and captions. 
In the next subsection, we provide the background necessary 
to understand how text is represented in most modern recurrent net applications. 
We then explain the commonly reported evaluation metrics.

\subsection{Representations of natural language inputs and outputs}

When words are output at each time step,
generally the output consists of a softmax vector $\boldsymbol{y}^{(t)} \in \mathbbm{R}^{K}$ 
where $K$ is the size of the vocabulary.
A softmax layer is an element-wise logistic function
that is normalized so that all of its components sum to one. 
Intuitively, these outputs correspond to the probabilities 
that each word is the correct output at that time step.

For application where an input consists of a sequence of words, 
typically the words are fed to the network one at a time in consecutive time steps.
In these cases, the simplest way to represent words is a \emph{one-hot} encoding,
using binary vectors with a length equal to the size of the vocabulary,
so ``1000" and ``0100" would represent the first and second words in the vocabulary respectively.
Such an encoding is discussed by \citet{elman1990finding} among others.
However, this encoding is inefficient, requiring as many bits as the vocabulary is large.
Further, it offers no direct way to capture different aspects 
of similarity between words in the encoding itself.
Thus it is common now to model words with a 
distributed representation using a \emph{meaning vector}.
In some cases, these meanings for words are learned given a large corpus of supervised data,
but it is more usual to initialize the \emph{meaning vectors} 
using an embedding based on word co-occurrence statistics.
Freely available code to produce word vectors from these statistics include
\emph{GloVe} \citep{pennington2014glove}, 
and \emph{word2vec} \citep{goldberg2014word2vec},
which implements a word embedding algorithm from \citet{mikolov2013efficient}.

Distributed representations for textual data were described by \citet{hinton1986learning},
used extensively for natural language by \citet{bengio2003neural},
and more recently brought to wider attention in the deep learning community 
in a number of papers describing recursive auto-encoder (RAE) networks 
\citep{socher2010learning, socher2011dynamic, socher2011parsing, socher2011semi}.
For clarity we point out that these \emph{recursive} networks
are \emph{not} recurrent neural networks,
and in the present survey the abbreviation RNN always means {recurrent neural network}.
While they are distinct approaches, recurrent and recursive neural networks have important features in common,
namely that they both involve extensive weight tying and are both trained end-to-end via backpropagation.

In many experiments with recurrent neural networks 
\citep{elman1990finding, sutskever2011generating, zaremba2014learning}, 
input is fed in one character at a time, and output generated one character at a time,
as opposed to one word at a time.
While the output is nearly always a softmax layer,
many papers omit details of how they represent single-character inputs.
It seems reasonable to infer that characters are encoded with a one-hot encoding. 
We know of no cases of paper using a distributed representation at the single-character level.

\subsection{Evaluation methodology}

A serious obstacle to training systems well to output variable length sequences of words 
is the flaws of the available performance metrics.
In the case of captioning or translation,
there maybe be multiple correct translations. 
Further, a labeled dataset may contain multiple \emph{reference translations}
for each example.
Comparing against such a gold standard is more fraught 
than applying standard performance measure to binary classification problems.

One commonly used metric for structured natural language output with multiple references is $\textit{BLEU}$ score.
Developed in 2002, 
$\textit{BLEU}$ score is related to modified unigram precision \citep{papineni2002bleu}.
It is the geometric mean of the $n$-gram precisions 
for all values of $n$ between $1$ and some upper limit $N$.
In practice, $4$ is a typical value for $N$, shown to maximize agreement with human raters. 
Because precision can be made high by offering excessively short translations, 
the $\textit{BLEU}$ score includes a brevity penalty $B$.
Where $c$ is average the length of the candidate translations and $r$ the average length of the reference translations,
the brevity penalty is 
$$ 
B = 
\begin{cases} 
1 			&\mbox{if } 	c > r \\ 
e^{(1-r/c)} 	& \mbox{if } 	c \leq r 
\end{cases}.
$$
Then the $\textit{BLEU}$ score is
$$
\textit{BLEU} = B \cdot \exp \left( \frac{1}{N} \sum_{n=1}^{N}  \log p_n  \right)
$$
where $p_n$ is the modified $n$-gram precision,
which is the number of $n$-grams in the candidate translation
that occur in any of the reference translations,
divided by the total number of $n$-grams in the candidate translation.
This is called \emph{modified} precision because it is an adaptation of precision to the case of multiple references.

$\textit{BLEU}$ scores are commonly used in recent papers to evaluate both translation and captioning systems.
While $\textit{BLEU}$ score does appear highly correlated with human judgments,
there is no guarantee that any given translation with a higher $\textit{BLEU}$ score 
is superior to another which receives a lower $\textit{BLEU}$ score.
In fact, while $\textit{BLEU}$ scores tend to be correlated with human judgement across large sets of translations,
they are not accurate predictors of human judgement at the single sentence level.

$\textit{METEOR}$ is an alternative metric intended to overcome the weaknesses 
of the $\textit{BLEU}$ score \citep{banerjee2005meteor}.
$\textit{METEOR}$ is based on explicit word to word matches 
between candidates and reference sentences. 
When multiple references exist, the best score is used.
Unlike $\textit{BLEU}$, $\textit{METEOR}$ exploits known synonyms and stemming.
The first step is to compute an F-score
$$
F_{\alpha} = \frac{P \cdot R}{\alpha \cdot P + (1-\alpha) \cdot R}
$$
based on single word matches where $P$ is the precision and $R$ is the recall.
The next step is to calculate a fragmentation penalty $M \propto c/m$ 
where $c$ is the smallest number of \emph{chunks} of consecutive words 
such that the words are adjacent in both the candidate and the reference, 
and $m$ is the total number of matched unigrams yielding the score.
Finally, the score is 
$$
\textit{METEOR} = (1 - M) \cdot F_\alpha.
$$
Empirically, this metric has been found to agree with human raters more than $\textit{BLEU}$ score.
However, $\textit{METEOR}$ is less straightforward to calculate than $\textit{BLEU}$. 
To replicate the $\textit{METEOR}$ score reported by another party, 
one must exactly replicate their stemming and synonym matching,
as well as the calculations. 
Both metrics rely upon having the exact same set of reference translations. 

Even in the straightforward case of binary classification, without sequential dependencies, 
commonly used performance metrics like F1 
give rise to optimal thresholding strategies 
which may not accord with intuition about what should constitute good performance \citep{lipton2014optimal}.
Along the same lines, given that performance metrics such as the ones above 
are weak proxies for true objectives, 
it may be difficult to distinguish between systems which are truly stronger 
and those which most overfit the performance metrics in use.

\subsection{Natural language translation}

Translation of text is a fundamental problem in machine learning 
that resists solutions with shallow methods.
Some tasks, like document classification, 
can be performed successfully with a bag-of-words representation that ignores word order. 
But word order is essential in translation. 
The sentences {``Scientist killed by raging virus"} 
and {``Virus killed by raging scientist"} have identical bag-of-words representations.
	
\begin{figure}
  \center
  \includegraphics[width=1\linewidth]{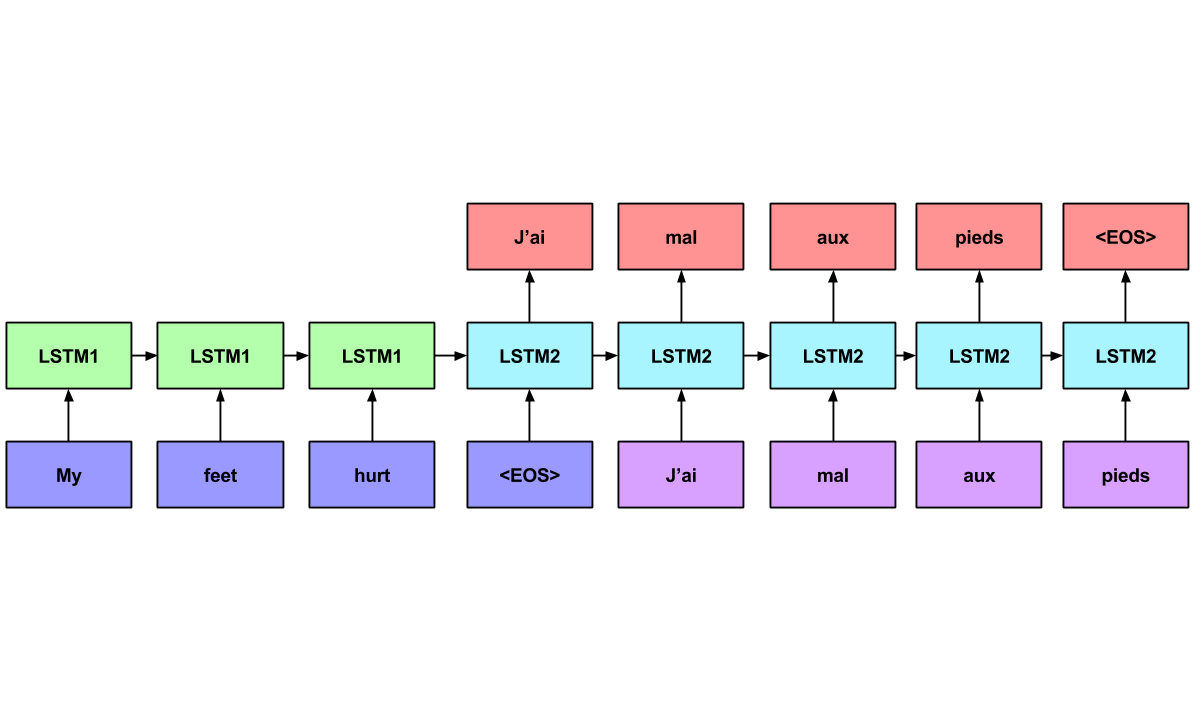}
  \caption{Sequence to sequence LSTM model of \citet{sutskever2014sequence}. 
  The network consists of an encoding model (first LSTM) and a decoding model (second LSTM). 
The input blocks (blue and purple) correspond to word vectors, 
which are fully connected to the corresponding hidden state. 
Red nodes are softmax outputs. Weights are tied among all encoding steps and among all decoding time steps. }
  \label{fig:sequence-to-sequence}
\end{figure}

\citet{sutskever2014sequence} present a translation model using two multilayered LSTMs 
that demonstrates impressive performance translating from English to French.
The first LSTM is used for \emph{encoding} an input phrase from the source language 
and the second LSTM for \emph{decoding} the output phrase in the target language.
The model works according to the following procedure (Figure~\ref{fig:sequence-to-sequence}): 
\begin{itemize}
	\item 
	The source phrase is fed to the {encoding} LSTM one word at a time,
	which does not output anything. 
	The authors found that significantly better results are achieved 
	when the input sentence is fed into the network in reverse order.
	\item 
	When the end of the phrase is reached, a special symbol that indicates 
	the beginning of the output sentence is sent to the {decoding} LSTM. 
	Additionally, the {decoding} LSTM receives as input the final state of the first LSTM. 
	The second LSTM outputs softmax probabilities over the vocabulary at each time step.
	\item
	At inference time, beam search is used to choose the most likely words from the distribution at each time step, 
	running the second LSTM until the end-of-sentence (\emph{EOS}) token is reached.
\end{itemize}

For training, the true inputs are fed to the encoder,
the true translation is fed to the decoder,
and loss is propagated back from the outputs of the decoder across the entire sequence to sequence model.
The network is trained to maximize the likelihood of the correct translation of each sentence in the training set. 
At inference time, a left to right beam search is used to determine which words to output. 
A few among the most likely next words are chosen for expansion after each time step.
The beam search ends when the network outputs an end-of-sentence (\emph{EOS}) token.
\citet{sutskever2014sequence}  train the model using stochastic gradient descent without momentum, 
halving the learning rate twice per epoch, after the first five epochs.
The approach achieves a $\textit{BLEU}$ score of 34.81, 
outperforming the best previous neural network NLP systems, 
and matching the best published results for non-neural network approaches, 
including systems that have explicitly programmed domain expertise.
When their system is used to rerank candidate translations from another system,
it achieves a $\textit{BLEU}$ score of 36.5.

The implementation which achieved these results involved eight GPUS. 
Nevertheless, training took 10 days to complete.
One GPU was assigned to each layer of the LSTM,
and an additional four GPUs were used simply to calculate softmax.
The implementation was coded in C++,
and each hidden layer of the LSTM contained 1000 nodes.
The input vocabulary contained 160,000 words and the output vocabulary contained 80,000 words.
Weights were initialized uniformly randomly in the range between $-0.08$ and $0.08$.

Another RNN approach to language translation is presented by \citet{auli2013joint}.
Their RNN model uses the word embeddings of Mikolov 
and a lattice representation of the decoder output 
to facilitate search over the space of possible translations.
In the lattice, each node corresponds to a sequence of words.
They report a $\textit{BLEU}$ score of 28.5 on French-English translation tasks.
Both papers provide results on similar datasets but 
\citet{sutskever2014sequence} only report on English to French translation 
while \citet{auli2013joint} only report on French to English translation,
so it is not possible to compare the performance of the two models. 

\subsection{Image captioning}

Recently, recurrent neural networks have been used successfully 
for generating sentences that describe photographs
\citep{vinyals2015show, karpathy2014deep, mao2014deep}.
In this task, a training set consists of input images $\boldsymbol{x}$ and target captions $\boldsymbol{y}$. 
Given a large set of image-caption pairs, a model is trained to predict the appropriate caption for an image.

\citet{vinyals2015show} follow up on the success in language to language translation 
by considering captioning as a case of image to language translation.
Instead of both encoding and decoding with LSTMs,
they introduce the idea of encoding an image with a convolutional neural network, 
and then decoding it with an LSTM. 
\citet{mao2014deep} independently developed a similar RNN image captioning network, 
and achieved then state-of-the-art results 
on the Pascal, Flickr30K, and COCO datasets.

\citet{karpathy2014deep} follows on this work, 
using a convolutional network to encode images 
together with a bidirectional network attention mechanism
and standard RNN to decode captions, 
using word2vec embeddings as word representations. 
They consider both full-image captioning and a model 
that captures correspondences between image regions and text snippets. 
At inference time, their procedure resembles the one described by \citet{sutskever2014sequence}, 
where sentences are decoded one word at a time. 
The most probable word is chosen and fed to the network at the next time step. 
This process is repeated until an \emph{EOS} token is produced.

To convey a sense of the scale of these problems,
\citet{karpathy2014deep} focus on three datasets of captioned images:
Flickr8K, Flickr30K, and COCO,
of size 50MB (8000 images), 200MB (30,000 images), and 750MB (328,000 images) respectively.
The implementation uses the Caffe library \citep{jia2014caffe},
and the convolutional network is pretrained on ImageNet data.
In a revised version, the authors report that LSTMs outperform simpler RNNs
and that learning word representations from random initializations
is often preferable to {word2vec} embeddings.
As an explanation, they say that {word2vec} embeddings
may cluster words like colors together in the embedding space,
which can be not suitable for visual descriptions of images.

\subsection{Further applications}

Handwriting recognition is an application area
where bidirectional LSTMs have been used to achieve state of the art results.
In work by \citet{liwicki2007novel} and \citet{graves2009novel},
data is collected from an interactive whiteboard.
Sensors record the $(x,y)$ coordinates of the pen at regularly sampled time steps.
In the more recent paper, they use a bidirectional LSTM model,
outperforming an HMM model by achieving $81.5\%$
word-level accuracy, compared to $70.1\%$ for the HMM.

In the last year, a number of papers have emerged
that extend the success of recurrent networks
for translation and image captioning to new domains.
Among the most interesting of these applications 
are unsupervised video encoding \citep{srivastava2015unsupervised}, 
video captioning \citep{venugopalan2015sequence},
and program execution \citep{zaremba2014learning}.
\citet{venugopalan2015sequence}~demonstrate a sequence to sequence architecture 
that encodes frames from a video and decode words.
At each time step the input to the encoding LSTM 
is the topmost hidden layer of a convolutional neural network.
At decoding time, the network outputs probabilities over the vocabulary at each time step.

\citet{zaremba2014learning} experiment with networks which read 
computer programs one character at a time and predict their output.
They focus on programs which output integers
and find that for simple programs,
including adding two nine-digit numbers, 
their network, which uses LSTM cells in several stacked hidden layers
and makes a single left to right pass through the program,
can predict the output with 99\% accuracy.

\section{Discussion}

Over the past thirty years, recurrent neural networks have gone from 
models primarily of interest for cognitive modeling and computational neuroscience,
to powerful and practical tools for large-scale supervised learning from sequences.
This progress owes to advances in model architectures, 
training algorithms, and parallel computing.
Recurrent networks are especially interesting because they overcome 
many of the restrictions placed on input and output data by traditional machine learning approaches.
With recurrent networks, the assumption of independence between consecutive examples is broken,
and hence also the assumption of fixed-dimension inputs and outputs.

While LSTMs and BRNNs have set records in accuracy on many tasks in recent years,
it is noteworthy that advances come from novel architectures 
rather than from fundamentally novel algorithms.
Therefore, automating exploration of the space of possible models, 
for example via genetic algorithms or a Markov chain Monte Carlo approach,
could be promising.
Neural networks offer a wide range of transferable and combinable techniques. 
New activation functions, training procedures, initializations procedures, etc.~are generally transferable across networks and tasks,
often conferring similar benefits.
As the number of such techniques grows, the practicality of testing all combinations diminishes.
It seems reasonable to infer that as a community,
neural network researchers are exploring the space of model architectures and configurations
much as a genetic algorithm might,
mixing and matching techniques,
with a fitness function in the form of evaluation metrics on major datasets of interest.

This inference suggests two corollaries. 
First, as just stated, 
this body of research could benefit from automated procedures 
to explore the space of models. 
Second, as we build systems designed to perform more complex tasks,
we would benefit from improved fitness functions. 
A \textit{BLEU} score inspires less confidence than the accuracy reported on a binary classification task.
To this end, when possible, it seems prudent to individually test techniques 
first with classic feedforward networks on datasets with established benchmarks 
before applying them to recurrent networks in settings with less reliable evaluation criteria.

Lastly, the rapid success of recurrent neural networks 
on natural language tasks leads us to believe that extensions of this work to longer texts would be fruitful.
Additionally, we imagine that dialogue systems could be built 
along similar principles to the architectures used for translation, 
encoding prompts and generating responses, 
while retaining the entirety of conversation history as contextual information.

\section{Acknowledgements}
The first author's research is funded by generous support from the Division of Biomedical Informatics at UCSD, 
via a training grant from the National Library of Medicine.
This review has benefited from insightful comments from 
Vineet Bafna, Julian McAuley, Balakrishnan Narayanaswamy,
Stefanos Poulis, Lawrence Saul, Zhuowen Tu, and Sharad Vikram.


%
%
%
%
%
%

\bibliography{rnn_jmlr}
\end{document}